\documentclass[letterpaper, 10 pt, conference]{ieeeconf}  

\IEEEoverridecommandlockouts                              

\makeatletter
\let\NAT@parse\undefined
\makeatother

\usepackage[nospace]{cite}
\usepackage{amsmath,amssymb,amsfonts}
\usepackage{algorithmic}
\usepackage{graphicx}
\usepackage{textcomp}
\usepackage[dvipsnames]{xcolor}
\usepackage{multirow}
\usepackage{makecell}
\usepackage{tikz}
\usepackage{color, colortbl}
\usepackage{xcolor}
\usepackage{url}
\definecolor{LightBlue}{RGB}{212, 250, 252} 
\usepackage[flushleft]{threeparttable}
\usepackage{booktabs}
\usetikzlibrary{arrows.meta}
\let\labelindent\relax
\usepackage{enumitem}
\definecolor{LightBlue}{RGB}{212, 250, 252}
\usepackage{caption}
\usepackage{subcaption}
\usepackage{tabulary}
\usepackage{pifont}      
\usepackage{siunitx}

\usepackage[hidelinks]{hyperref}

\def\mycircle[#1]{\tikz\draw[#1,fill=#1] (0,0) circle (0.125cm);}
\newcommand{\cmark}{\ding{51}}

\def\arrowright[#1]{\begin{tikzpicture}
  \draw[#1, -{Triangle[width = 5pt, length = 2pt]}, line width = 2pt] (0.0, 0.0) -- (0.25, 0.0);\path (current bounding box.south west) +(0,-0.06);
\end{tikzpicture}}

\def\arrowleft[#1]{\begin{tikzpicture}
  \draw[#1, -{Triangle[width = 5pt, length = 2pt]}, line width = 2pt] (0.25, 0.0) -- (0.0, 0.0);\path (current bounding box.south west) +(0,-0.06);

\end{tikzpicture}}

\def\arrowup[#1]{\begin{tikzpicture}
  \draw[#1, -{Triangle[width = 5pt, length = 2pt]}, line width = 2pt, rotate=270] (0.25, 0.0) -- (0.0, 0.0);
\end{tikzpicture}}

\def\arrowstraightleft[#1]{\begin{tikzpicture}
\draw[#1, -{Triangle[width = 5pt, length = 2pt]}, line width = 2pt, rotate=270] (0.25, 0.0) -- (0.0, 0.0);
  \draw[#1, -{Triangle[width = 5pt, length = 2pt]}, line width = 2pt] (0.0, -0.17) -- (-0.15, -0.17);
\end{tikzpicture}}

\def\arrowstraightright[#1]{\begin{tikzpicture}
    \draw[#1, -{Triangle[width = 5pt, length = 2pt]}, line width = 2pt, rotate=270] (0.25, 0.0) -- (0.0, 0.0);

  \draw[#1, -{Triangle[width = 5pt, length = 2pt]}, line width = 2pt] (0.0, -0.17) -- (0.15, -0.17);

\end{tikzpicture}}

\definecolor{af-white}{rgb}{0.95, 0.95, 0.96}
\definecolor{Pistachio}{RGB}{140, 212, 126} 
\definecolor{CrayolaYellow}{RGB}{248, 214, 109}
\definecolor{PastelRed}{RGB}{255, 105, 97}
\definecolor{PastelOrange}{RGB}{255, 181, 76}
\definecolor{SoftCharcoal}{RGB}{66,66, 66}
\title{\LARGE \bf The ATLAS of Traffic Lights: A Reliable Perception Framework for Autonomous Driving}

\author{Rupert Polley$^{\ast1}$, Nikolai Polley$^{\ast2}$, Dominik Heid$^{1}$, Marc Heinrich$^{1}$, Sven Ochs$^{1}$, and J. Marius Zöllner$^{1,2}$
\thanks{$^\ast$ Equal contribution.}%
\thanks{$^{1}$ FZI Research Center for Information Technology, \mbox{Technical} Cognitive Systems, Haid-und-Neu Str. 10-14, Karlsruhe, \mbox{Germany}.
	{\tt\small \{surname\}@fzi.de}.}%
\thanks{$^{2}$ Karlsruhe Institute of Technology (KIT), Research Group Applied Technical-Cognitive Systems, Kaiserstr. 12, Karlsruhe, Germany. {\tt\small \{prename.surname\}@kit.edu}.}
}

\usepackage{eso-pic}
\newcommand{\mycopyrighttext}{%
  \footnotesize
  \noindent
  \textcopyright~2025 IEEE. Personal use of this material is permitted.
  Permission from IEEE must be obtained for all other uses, in any current
  or future media, including reprinting/republishing this material for
  advertising or promotional purposes, creating new collective works,
  for resale or redistribution to servers or lists, or reuse of any
  copyrighted component of this work in other works.\\
  IEEE 36th Intelligent Vehicles Symposium (IV 2025) - 22-25 June, 2025.
}

\AtBeginDocument{%
  \AddToShipoutPicture*{%
    \AtPageUpperLeft{%
      \put(55, -40){
        \parbox{\dimexpr\textwidth-4pt\relax}{\raggedright \mycopyrighttext}
      }
    }
  }
}

\begin{document}

\maketitle


\thispagestyle{empty}
\pagestyle{empty}

\begin{abstract} Traffic light perception is an essential component of the camera-based perception system for autonomous vehicles, enabling accurate detection and interpretation of traffic lights to ensure safe navigation through complex urban environments. In this work, we propose a modularized perception framework that integrates state-of-the-art detection models with a novel real-time association and decision framework, enabling seamless deployment into an autonomous driving stack. To address the limitations of existing public datasets, we introduce the \texttt{ATLAS} dataset, which provides comprehensive annotations of traffic light states and pictograms across diverse environmental conditions and camera setups. This dataset is publicly available at \url{https://url.fzi.de/ATLAS}. We train and evaluate several state-of-the-art traffic light detection architectures on \texttt{ATLAS}, demonstrating significant performance improvements in both accuracy and robustness. Finally, we evaluate the framework in real-world scenarios by deploying it in an autonomous vehicle to make decisions at traffic light-controlled intersections, highlighting its reliability and effectiveness for real-time operation.

\end{abstract}

\section{Introduction}
Perception of traffic lights plays a pivotal role in ensuring the safe navigation of urban environments for autonomous driving (AD). To operate reliably, autonomous vehicles must not only detect and classify traffic lights accurately but also interpret their relevance to the vehicle's current context and programmed trajectory. Complex intersections, occlusions, and environmental conditions such as rain or nighttime visibility remain a challenge. Unlike other perception tasks, such as object detection, where LiDAR can complement vision-based approaches, traffic light recognition primarily relies on real-time camera-based perception. 
While Vehicle-to-Everything (V2X) communication has the potential to provide traffic light state information, its deployment remains sparse, making vision-based detection the only widely available method. Public datasets traditionally used for training traffic light detection models often fail to comprehensively cover all traffic light states, particularly lane-specific direction indicators (pictograms), and lack annotations for diverse environmental conditions, such as heavy rain. Moreover, all public datasets provide only a single field of view (FOV) camera setup, limiting their applicability to real-world autonomous systems, which require multiple cameras with varying FOVs to detect traffic lights at different distances.

To tackle these challenges, we contribute the following: 
\begin{itemize}
\item     We introduce the \texttt{ATLAS} dataset, which remedies key deficiencies in existing datasets by providing annotations for a wide range of traffic light states, pictograms, and environmental conditions across multi-camera setups with different FOVs.
\item     We conduct an extensive evaluation of state-of-the-art detection models trained on \texttt{ATLAS}, achieving significant improvements in detection accuracy and generalization.
\item We propose novel real-time association and decision-making modules, enabling robust and stable traffic light detection in complex urban environments.
\item We validate the entire traffic light perception framework by integrating it into an autonomous vehicle, demonstrating its high accuracy and robustness for decision-making at traffic light-controlled intersections.
\end{itemize}

\begin{figure}[t]
\centering

\begin{subfigure}[t]{1\linewidth}
  \includegraphics[width=1.0\textwidth]{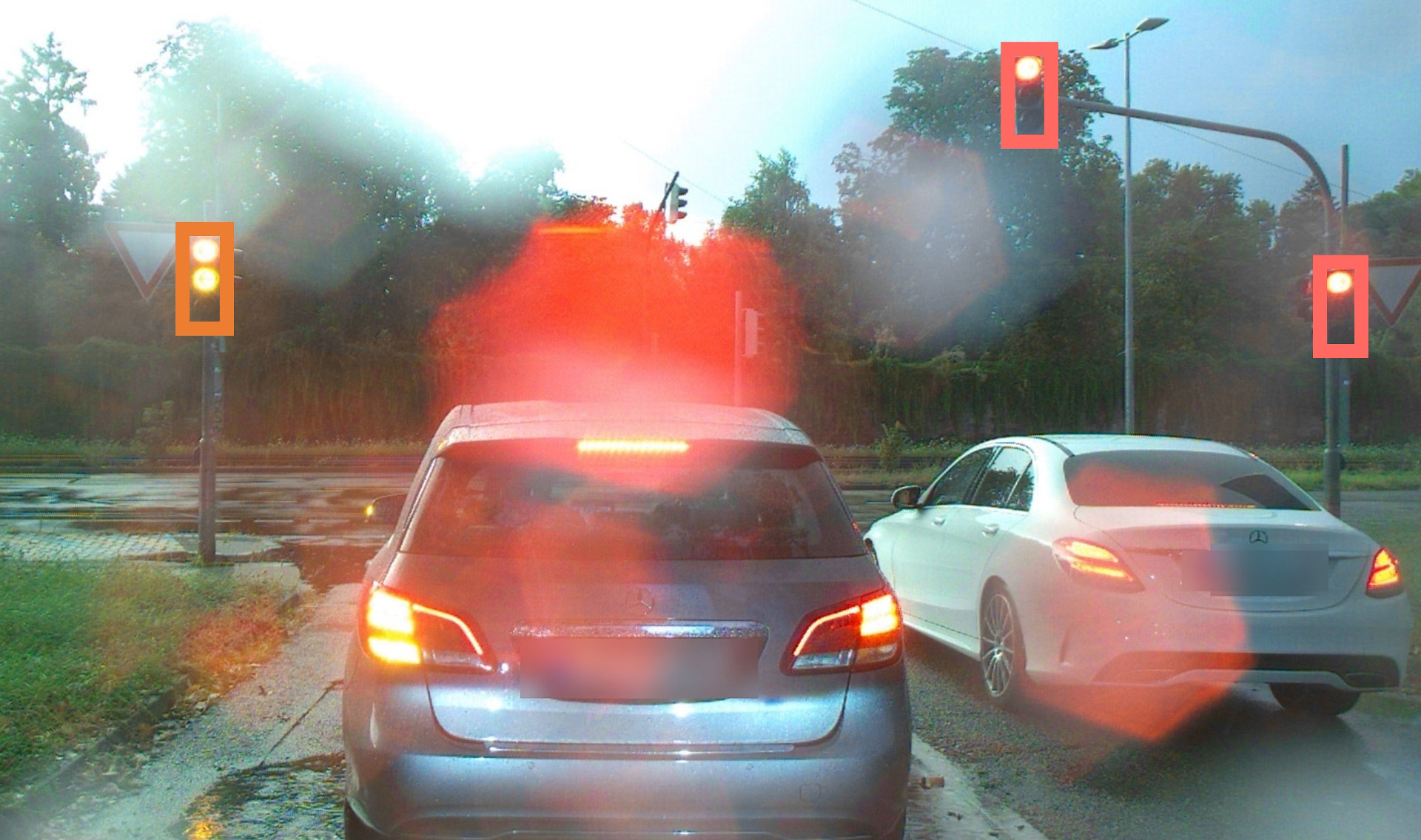}
\end{subfigure}

\vspace{0.2cm}
\begin{subfigure}[t]{0.49\linewidth}
    \includegraphics[width=1.0\textwidth]{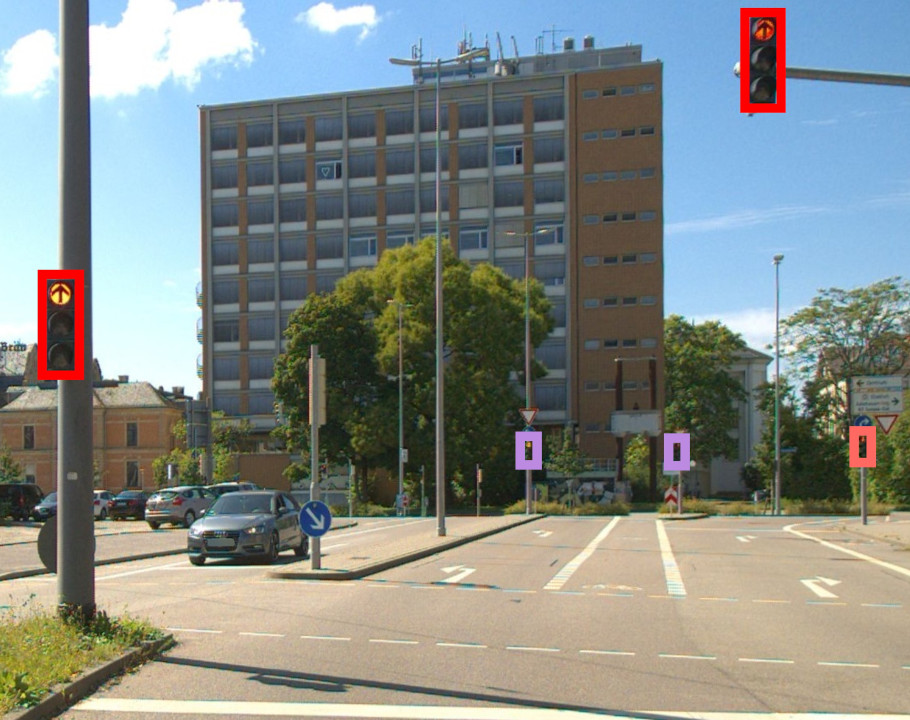}
\end{subfigure}
\hfill
\begin{subfigure}[t]{0.49\linewidth}
  \includegraphics[width=1.0\textwidth]{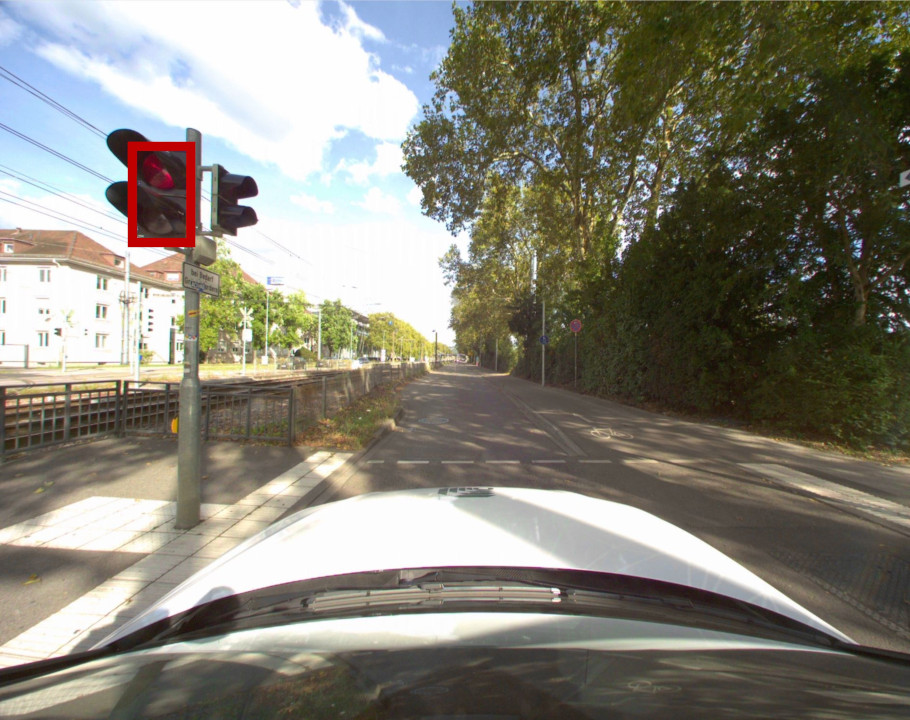}
\end{subfigure}
\caption{Annotated images from \texttt{ATLAS}. The top image captures a scene recorded in rain, featuring a traffic light in the \textit{red-yellow} state. The bottom left image shows multiple \textit{red} traffic lights with distinct pictograms differentiated by color. The bottom right image demonstrates the necessity of a wide FOV, as the on-demand traffic light is not visible within a standard camera's field of view.}
\label{fig:ATLAS_images}
\end{figure}

\newpage
\section{Related Work}

\subsection{Traffic Light Detectors}
In the early stages of autonomous driving research, traffic light detectors primarily relied on calculating image features like the shape, color, and brightness characteristics of traffic lights to identify their presence and position within the visual field~\cite{chung2002vision, de2009real, lindner2004robust, nienhuser2010visual}. In more recent years, research in this field has increasingly shifted towards learned approaches, particularly those utilizing convolutional neural networks (CNNs)~\cite{pavlitska2023traffic}. Here, most often, 2D-convolutional network object detectors are learned to predict rectangular bounding boxes encompassing traffic lights and to classify their corresponding states. Possatti et al.~\cite{possatti2019traffic} train an object detector for the classes stop/go. In inference, they use the most recent predicted bounding box, which leads to rapid flickering in the predictions and rapidly changing autonomous driving instructions. 
Chen et al.~\cite{chen2024traffic} propose an ensemble of object detectors, detecting the individual bulbs of traffic lights.
In prior work, we proposed a modification of the  YOLOv8~\cite{jocher2023yolov8} architecture, increasing performance for far-away small traffic lights~\cite{tldready}. 
\subsection{Datasets for Traffic Light Detection}

\begin{table*}[t]
\caption{Public datasets evaluated for inclusion of traffic light states (\textit{green}, \textit{red}, \textit{yellow}, \textit{red-yellow}) and pictograms (\textit{circle}, \textit{off}, \textit{straight}, \textit{left}, \textit{right}, \textit{straight-left}, \textit{straight-right}). If less than 30 annotations exist, we show the total number of annotations for this class. \texttt{ATLAS} is the only dataset containing the \textit{straight-right} pictogram. }
\label{tab:classes}

\setlength{\extrarowheight}{4.5pt}
\resizebox{1.0\linewidth}{!}{

\begin{tabular}{| r|c|c|c|c|c|c|c|c|c|c|c|c|c|c|c|c|c|c|c|c|c|c|c|c|c|}
\hline
\textbf{Dataset} & \mycircle[Pistachio] & \mycircle[PastelRed] & \mycircle[CrayolaYellow] & \mycircle[PastelOrange] & \mycircle[SoftCharcoal] & \arrowup[Pistachio] & \arrowup[PastelRed] & \arrowup[CrayolaYellow] & \arrowup[PastelOrange] & \arrowleft[Pistachio] & \arrowleft[PastelRed] & \arrowleft[CrayolaYellow] & \arrowleft[PastelOrange] & \arrowright[Pistachio] & \arrowright[PastelRed] & \arrowright[CrayolaYellow] & \arrowright[PastelOrange] & \arrowstraightleft[Pistachio] & \arrowstraightleft[PastelRed] & \arrowstraightleft[CrayolaYellow] & \arrowstraightleft[PastelOrange] & \arrowstraightright[Pistachio] & \arrowstraightright[PastelRed] & \arrowstraightright[CrayolaYellow] & \arrowstraightright[PastelOrange]\\\hline

\texttt{BSTLD} & \cmark \cellcolor{af-white} & \cmark \cellcolor{af-white} & \cmark \cellcolor{af-white} & & \cmark \cellcolor{af-white} & & & & & & & & & & & & & & & & &&&&\\ \hline

\texttt{LISA} & \cmark \cellcolor{af-white} & \cmark \cellcolor{af-white} & \cmark \cellcolor{af-white} & & & \cmark \cellcolor{af-white} & & & & \cmark \cellcolor{af-white} & \cmark \cellcolor{af-white} & \cmark \cellcolor{af-white} & & & & & & & & & &&&&\\ \hline

\texttt{DTLD} & \cmark \cellcolor{af-white} & \cmark \cellcolor{af-white} & \cmark \cellcolor{af-white} & \cmark \cellcolor{af-white} & \cmark \cellcolor{af-white} & \cmark \cellcolor{af-white} & \cmark \cellcolor{af-white} & \cmark \cellcolor{af-white} & \cmark \cellcolor{af-white} & \cmark \cellcolor{af-white} & \cmark \cellcolor{af-white} & \cmark \cellcolor{af-white} & \cmark \cellcolor{af-white} & \cmark \cellcolor{af-white} & \cmark \cellcolor{af-white} & \cmark \cellcolor{af-white} & \cellcolor{af-white}26& \cmark \cellcolor{af-white} & \cellcolor{af-white} 9& \cellcolor{af-white}2& &&&& \\ \hline

\texttt{\textbf{ATLAS}} & \cmark \cellcolor{af-white} & \cmark \cellcolor{af-white} & \cmark \cellcolor{af-white} & \cmark \cellcolor{af-white} & \cmark \cellcolor{af-white} & \cmark \cellcolor{af-white} & \cmark \cellcolor{af-white} & \cmark \cellcolor{af-white} & \cmark \cellcolor{af-white} & \cmark \cellcolor{af-white} & \cmark \cellcolor{af-white} & \cmark \cellcolor{af-white} & \cmark \cellcolor{af-white} & \cmark \cellcolor{af-white} & \cmark \cellcolor{af-white} & \cmark \cellcolor{af-white} & \cmark \cellcolor{af-white} & \cmark \cellcolor{af-white} & \cmark \cellcolor{af-white} & \cmark \cellcolor{af-white} &\cmark \cellcolor{af-white} &\cmark \cellcolor{af-white} &\cmark \cellcolor{af-white} &\cmark \cellcolor{af-white} &\cellcolor{af-white}{\centering\arraybackslash 20} \\ \hline
\end{tabular}
}
\end{table*}

Large-scale autonomous driving datasets, such as KITTI~\cite{Geiger2012CVPR} and nuScenes~\cite{nuscenes}, do not include annotations for traffic light states or pictograms in camera frames. Consequently, several specialized datasets have been proposed to address this limitation. Below, we provide a concise overview of publicly available datasets containing more than 10,000 annotated images and meet the requirements for training learned object detectors. \texttt{LaRa}~\cite{LARA} contains 11,000 images but does not annotate pictograms and its resolution of 640$\times$480 is quite low. \texttt{aiMotive}~\cite{kunsagi2024aimotive} comprises 50,000 images with automatically generated annotations, but visual inspection reveals a significant number of errors, particularly in state annotations. The Bosch Small Traffic Light Dataset \texttt{BSTLD}~\cite{behrendt2017deep} contains 13,000 images with a resolution of 1280$\times$720. In the training dataset, pictograms of the traffic lights, like arrows denoting corresponding driving lanes, are annotated, but in the test split, only four traffic light states \textit{red}, \textit{yellow}, \textit{green} and \textit{off} are annotated. \texttt{LISA}~\cite{jensen2016vision} is a large-scale dataset of 36,000 images containing state annotation and some pictograms. However, the dataset exhibits inconsistencies in annotation quality, as numerous traffic lights remain unlabeled, and the provided bounding boxes frequently fail to accurately encompass the traffic lights~\cite{fregin2018driveu}.

To our knowledge, the DriveU Traffic Light Dataset \texttt{DTLD}~\cite{fregin2018driveu} is the largest and most comprehensive public dataset with 40,000 annotated images including annotated pictograms and the \textit{red-yellow} traffic light state, which is common in Germany and occurs when a \textit{red} light transitions to \textit{green}. Table~\ref{tab:classes} illustrates the contained classes and pictograms of these datasets.

\vfill

\subsection{Association and Relevance Estimation}

Camera-based traffic light detection systems typically output the pixel coordinates and states of detected traffic lights. 
In complex traffic scenarios, such as multi-lane intersections, multiple traffic lights may be present, each controlling specific lanes and displaying distinct states.
Consequently, it is crucial to process the detection results to differentiate between traffic lights relevant to the ego vehicle and those that are not.

The methods of assigning relevancy vary significantly:
In one heuristic-based approach, the largest and topmost detected traffic light is always assigned as relevant~\cite{li2017traffic}. Our own tests reveal that this simplistic approach fails in a multitude of cases, such as where an overhead traffic light is incorrectly prioritized over a closer, lane-specific signal.

Langenberg et al.~\cite{langenberg2019deep} combine both image data with additional metadata, including information such as positions of traffic lights, arrows of the roads, and lane markings. A CNN-based classifier predicts the position of relevant traffic lights. While this method achieves higher accuracy than the previously mentioned heuristic approach, their approach is infeasible and untested for real-world driving, as the required metadata are human-annotated labels for each corresponding image, which are unavailable outside curated datasets.

A seminal work by Fairfield et al.~\cite{fairfield2011traffic} utilizes GPS and triangulation to create a mapping of traffic lights. Their method assumes that each traffic light is a fixed-size, three-bulb (\textit{red}-\textit{yellow}-\textit{green}) light and estimates the distance of a traffic light by counting the pixels of labeled traffic lights.  Over multiple frames, using least squares triangulation, they obtain a 3D position for each traffic light.
For real-world driving, the stored map positions are projected into the camera image at three times their original size. This enlarged region increases the likelihood that the traffic light is captured within, accounting for potential inaccuracies in real-time localization systems and map positions. For each of these regions, a blob-segmentation classifier predicts traffic light states. If the region is not classified, \textit{yellow} is assumed.

Possatti et al.~\cite{possatti2019traffic} combine prior maps with neural network-based detectors to identify relevant traffic lights. They create a map by driving a vehicle equipped with LiDAR sensors and annotating traffic lights in 3D space. The generated maps are route-specific, covering only traffic lights deemed relevant for the paths traversed during mapping. During inference, a 1.5-meter sphere is projected into the image space for each mapped traffic light. A CNN-based object detector then predicts bounding boxes within the image. Bounding boxes falling outside the projected spheres are discarded, while those within are assigned to the nearest sphere based on Euclidean distance. This approach is constrained by not inferring relevant traffic lights during autonomous driving, making it infeasible to navigate intersections differently from the specific paths driven during mapping. As a result, any deviation from the specific routes during mapping leads to the failure of their proposed model.

\newpage

\section{\texttt{ATLAS} Dataset}

In prior work~\cite{tldready}, we trained traffic light detection models on a variety of open-source datasets and noted that existing open-source datasets do not produce generalized classifiers viable for use in autonomous 
driving.
The reason is threefold:
\begin{enumerate}
    \item No existing public dataset contains all possible pictograms in all available traffic light states.
    \item No existing large-scale public dataset contains multiple annotated camera streams with varied FOVs.  
    \item No existing large-scale public datasets contain varied weather data.

\end{enumerate}
We observe that classifiers trained on existing public datasets produce numerous false positive predictions for environmental structures bearing slight resemblances to traffic lights. As these datasets do not contain critical pictograms like \textit{straight-right} or all states of \textit{straight-left} arrows (cf. Tab.~\ref{tab:classes}), they make it impossible for a trained model to predict these cases accurately. Similarly, we noticed severe performance degradation when applying models to high FOV cameras required for close-distance prediction. This issue is particularly significant for traffic lights mounted over the street, where the steep viewing angle prevents the model from reliably predicting the presence of the traffic light. 
We therefore propose the \textit{Applied Traffic Light Annotation Set} -  \texttt{ATLAS} - to mitigate some of these challenges and make it publicly available at \url{https://url.fzi.de/ATLAS}.

\begin{figure}[t]
\centering
\begin{subfigure}[t]{\linewidth}
    \includegraphics[width=\textwidth]{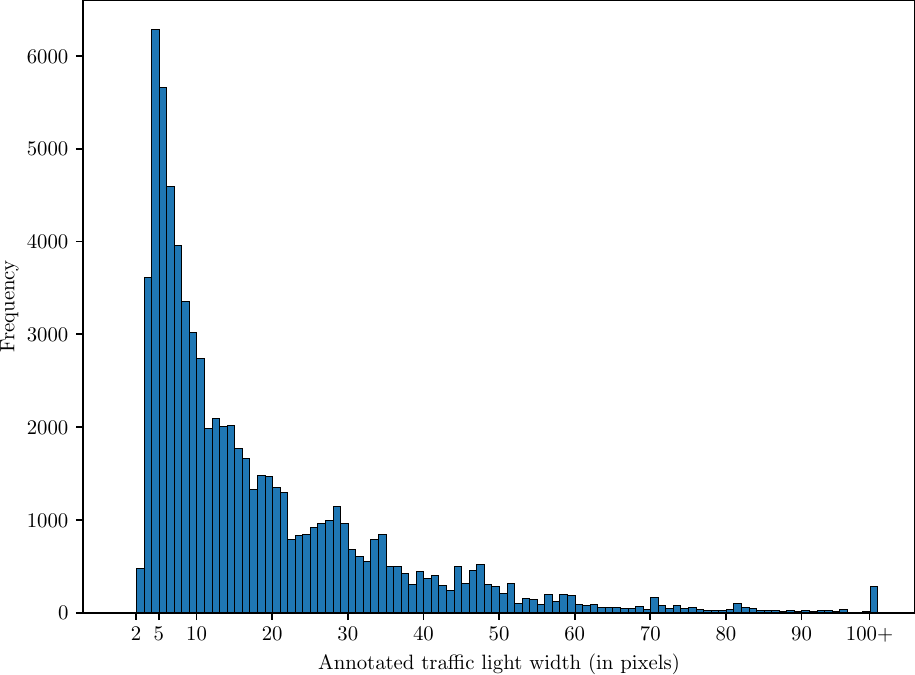}
\end{subfigure}
\caption{We start annotating traffic lights with two-pixel width. Most annotations are four pixels wide.}
    \label{fig:labelwidth}
\end{figure}

\begin{table}[b]
    \caption{The cameras used for dataset generation.}
    \label{tab:cameras}

  \begin{threeparttable}[b]
    \centering
\resizebox{1.0\linewidth}{!}{
    \begin{tabular}{r c c c}
    \toprule
    \textbf{Camera} & \textbf{FOV} [°] & \textbf{Resolution} & \textbf{Images} \\ \midrule
    Front-Medium & 61 $\times$ 39 & 1920 $\times$ 1200 & 25,158 \\
    Front-Tele & 31 $\times$ 20 & 1920 $\times$ 1200 & 5,109\\
    Front-Wide & 106 $\times$ 92 & 2592 $\times$ 2048 & 2,777 \\
                  \bottomrule
             \end{tabular}
}
    \end{threeparttable}%
\end{table}

\begin{figure}[t]
\centering
    \includegraphics[width=\linewidth]{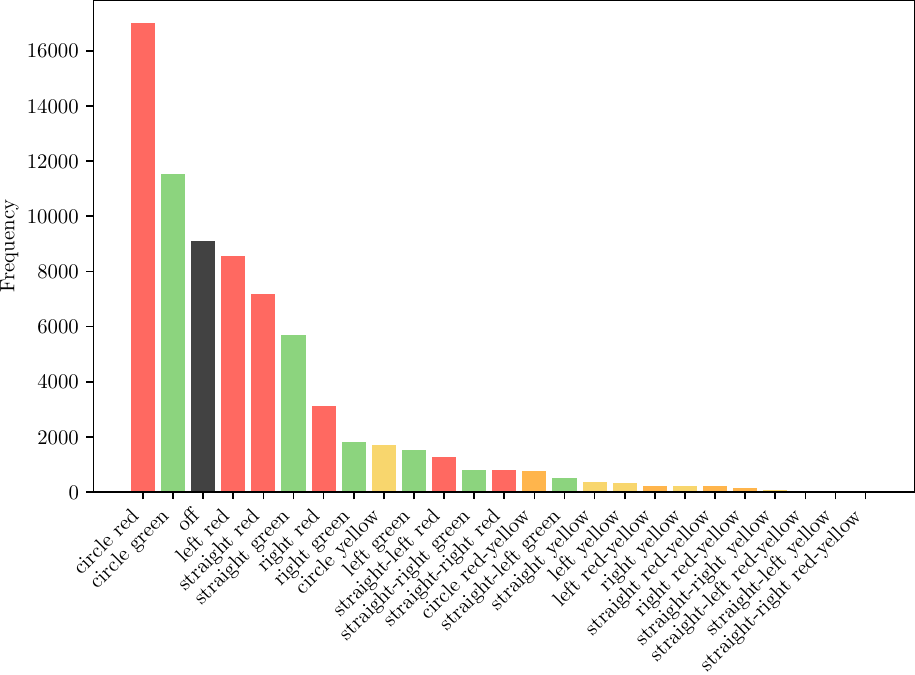}
    \caption{\texttt{ATLAS} contains twenty-five unique pictogram-state classes. As expected, the class imbalance is quite large.}
    \label{fig:classcounts}

\end{figure}

\subsection{Image Capturing}

Existing public datasets predominantly provide camera images from single- or stereo-camera setups with fixed focal lengths. However, real-world evaluations reveal that relying on a single fixed focal length is impractical. In European urban environments, traffic lights are positioned close to stopping lines, necessitating the use of short focal length lenses to ensure their visibility, see Figure~\ref{fig:ATLAS_images}. Conversely, the wide-angle images captured by short focal length lenses are suboptimal for detecting traffic lights at medium and long distances. The authors of \texttt{DTLD} also highlight this issue; however, they did not publish images and labels captured by wide-angle cameras.
In \texttt{ATLAS}, images from three separate cameras of our research vehicle CoCar NextGen~\cite{heinrich2024cocarnextgen} are annotated and presented.
Table~\ref{tab:cameras} details their configurations. The cameras are time-synchronized at 20~Hz, but to prevent labeling very similar images with little training advantages, we only annotate images at 2~Hz.

\subsection{Annotations}
We annotate all visible vehicle traffic lights. Traffic lights specifically for bicycles, trams, buses, or pedestrians are not annotated. Two-bulb on-demand traffic lights for vehicles without green bulbs are common in Germany and are therefore also annotated. 
For far-away traffic lights, we start labeling the moment their state becomes recognizable. This is usually when they are 2~-~4 pixels wide, which corresponds to an approximate distance of 130~meters for front-medium, 200~meters for front-tele, and 50~meters for front-wide. Figure~\ref{fig:labelwidth} illustrates the widths of all annotated traffic lights in all images.
For far-away traffic lights where only the state but not the pictogram is discernible in the image, we still label the pictogram. This is achieved by referring to later recorded images or utilizing Google Street View for traffic lights that were not captured up close. We use a self-supervised iterative approach for annotating traffic lights. We record data, utilize the current best model for initial predictions, manually correct these predictions, retrain the model on the enlarged dataset, and iterate on this process.
In the first iterations, the model, at this time mostly trained on \texttt{DTLD}, makes too many mistakes, so we remove these predictions and annotate from scratch. 
In later iterations, predictions improve significantly, and human labeling focuses primarily on tightening predicted bounding boxes to better-fit traffic light housings, as classification and localization are already highly accurate. 
For all images, we use a two-step human annotation process: one annotator refines model predictions by adding missing bounding boxes or correcting inaccurate ones. A second separate person reviews these annotations and, if needed, further corrects the initial human annotations. This ensures high annotation quality. We reduce the number of images recorded when the vehicle is stationary. As we want to increase the number of rare states, most notably \textit{yellow} and \textit{red-yellow}, we include all images in which these are visible, even if the vehicle is stationary. 
We notice that \texttt{DTLD} and \texttt{LISA} do not contain images during rain, and \texttt{BSTLD} contains only a few images with very light rain.  Consequently, we observed a severe degradation of our model performance during rain. To address this, we annotate and add 2,776 images of medium to heavy rain to our dataset (see Fig.~\ref{fig:ATLAS_images}).
Finally, we collect a large volume of data but only annotate and add images where the model makes specific errors, such as false positives, missed predictions (often at very long distances), or incorrect classifications (primarily of pictograms at long distances). 
In total, our dataset numbers 33,044 images with 72,998 annotated bounding boxes, with the number of each individual class illustrated in Fig.~\ref{fig:classcounts}.

\subsection{Anonymization}
We anonymize the dataset with DeepPrivacy2~\cite{hukkelas23DP2} and EgoBlur~\cite{raina2023egoblur}. Gaussian blur is applied to anonymize individuals and license plates in the dataset. To measure its impact, we train a traffic light detection model on anonymized data and test it on non-anonymized data. When compared to models trained on the original, non-anonymized dataset, there is no observable performance degradation. 

\newpage
\section{Traffic Light Perception Framework}

\begin{figure*}[t]
\begin{subfigure}[b]{\textwidth}
    \includegraphics[width=1\textwidth]{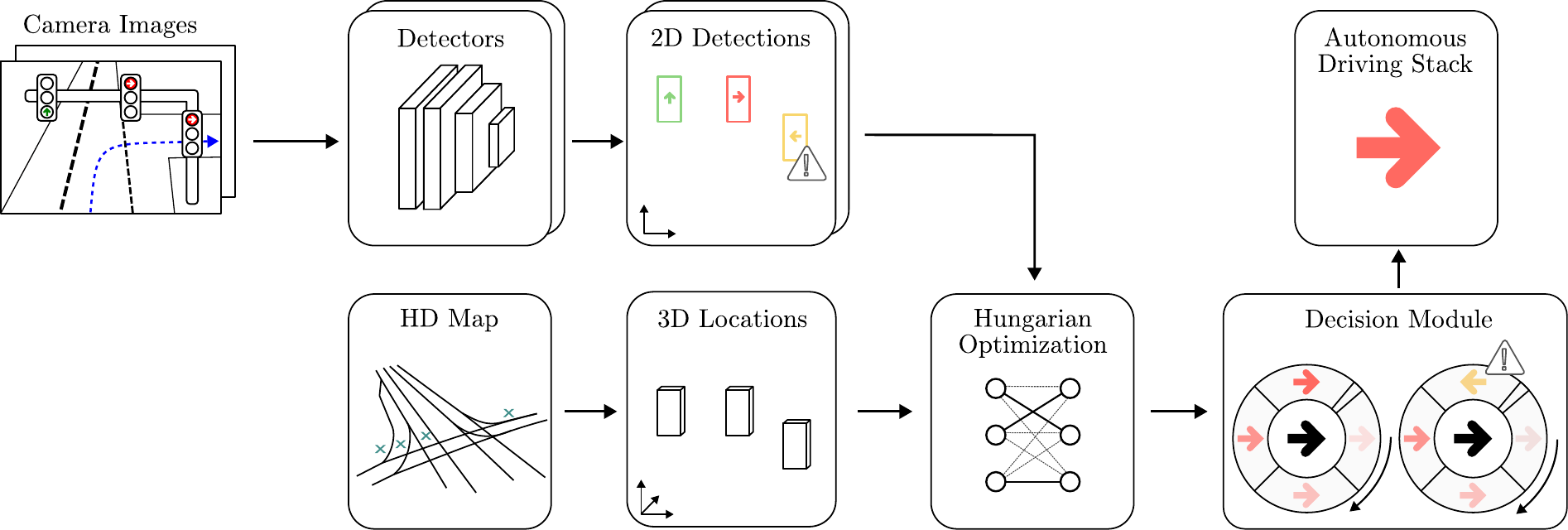}

\end{subfigure}
 \caption{Overview of our proposed traffic light perception framework. An autonomous vehicle approaches an intersection intending to make a right turn. Two detectors identify traffic lights in both camera streams. The detections are projected and associated with traffic light reference positions from the HD map using the Hungarian algorithm. Associated detections are stored in their respective circular buffers, and the system reasons over all traffic lights in a signal group to determine the final state. In this example, despite an incorrect state and pictogram prediction by the detector, the decision module correctly infers the true traffic light state.}
\label{fig:DecisionFramework}

\end{figure*}

We present a perception framework for detecting and responding to traffic lights with high accuracy and reliability in autonomous driving scenarios, illustrated in Fig~\ref{fig:DecisionFramework}. Our approach is structured with separate key modules, each addressing a critical aspect of the detection and decision-making method: 

\begin{itemize} 
\item Traffic Light Detection: Detecting individual traffic lights in camera images using object detectors.
\item Projection and Association: Projecting and associating the detected traffic lights with mapped data to ensure accurate localization and context-aware decision-making. 
\item Decision Module: Improving stability, reliability, and perception quality by introducing a ring buffer, enabling effective tracking and smoothing of associated detections. \end{itemize}

\subsection{Traffic Light Detection}
For detecting and classifying traffic lights, we re-use our developed modification of a generic YOLOv8 architecture~\cite{tldready}  and apply it to the more recent YOLOv9, YOLOv10 and YOLO11 architectures~\cite{wang2025yolov9, wang2024yolov10, yolo11_ultralytics}. The high VRAM requirements of the attention-based YOLOv12~\cite{tian2025yolov12} architecture, renders it unsuitable for high-resolution image training. For training, we combine the training splits of \texttt{ATLAS} and \texttt{DTLD} and test the model on both the \texttt{ATLAS} and \texttt{DTLD} test splits. We train the models using distributed data parallelism across three Nvidia H100 GPUs. 

\begin{table}[b]
        \caption{
    Evaluation of detectors trained on \texttt{DTLD} and \texttt{ATLAS}. Training on both datasets results in similar test performance on \texttt{DTLD} and substantially improves performance on \texttt{ATLAS}. 
    }
    \label{tab:model_eval}

    \resizebox{1.0\linewidth}{!}{

  \begin{threeparttable}[t]
    \centering
\begin{tabular}{ l l c c c c }
    \toprule
               \multirow{2}{*}{\textbf{Model}} 
        & \multirow{2}{*}{\textbf{Training dataset}}
      & \multicolumn{2}{c}{\textbf{Test DTLD}}
      & \multicolumn{2}{c}{\textbf{Test ATLAS}} \\
      \cmidrule(lr){3-4} \cmidrule(lr){5-6}
      & & \textbf{mAP50} & \textbf{mAP50-95}
        & \textbf{mAP50} & \textbf{mAP50-95} \\
  \midrule
         YOLOv8x & \texttt{DTLD} & 0.62 & 0.43 & 0.53 & 0.34 \\
         YOLOv9e & \texttt{DTLD} & 0.62 & 0.43 & 0.53 & 0.35 \\
         YOLOv10x & \texttt{DTLD} & 0.58 & 0.39 & 0.49 & 0.33 \\
         YOLO11x & \texttt{DTLD} & 0.64 & 0.46 & 0.53 & 0.34 \\
\\
         YOLOv8x& \texttt{DTLD} + \texttt{ATLAS} & 0.59 & 0.43 & 0.70 & 0.52 \\ 
         YOLOv9c&\texttt{DTLD} + \texttt{ATLAS} & 0.60 & 0.42 & 0.72 & 0.53 \\
         YOLOv9e&\texttt{DTLD} + \texttt{ATLAS} & 0.60 & 0.44 & 0.71 & 0.53 \\
        YOLOv10m& \texttt{DTLD} + \texttt{ATLAS} & 0.58 & 0.41 & 0.71 & 0.52 \\
         YOLOv10x& \texttt{DTLD} + \texttt{ATLAS} & 0.58 & 0.42 & 0.71 & 0.54 \\
           YOLO11m& \texttt{DTLD} + \texttt{ATLAS} & 0.64 & 0.46 & 0.69 & 0.52 \\ 
      YOLO11x& \texttt{DTLD} + \texttt{ATLAS} & \textbf{0.66} & \textbf{0.49} & \textbf{0.72} & \textbf{0.54} \\
                  \bottomrule
\end{tabular}
  
    \end{threeparttable}
    }
\end{table}

Table~\ref{tab:model_eval} presents the evaluation of models using the mean average precision mAP50 and mAP50-95 metrics. Previously trained \texttt{DTLD}-based models show subpar results when evaluated on this new \texttt{ATLAS} test dataset. The quality of all retrained models is quite similar, with YOLO11x achieving the highest scores on \texttt{ATLAS} and \texttt{DTLD}. 

For deployment, we utilize two instances of the same model, running concurrently on two separate camera streams.
One instance consistently processes images from the front-medium camera, which is the most suitable option for general scenarios. The second image stream alternates between the front-tele camera and the front-wide camera depending on the distance to the upcoming intersection. By default, the front-tele camera is used to detect distant traffic lights that may be too small or not visible in the front-medium camera stream. When the vehicle comes within 10 meters of the intersection, the second stream switches from the front-tele camera to the front-wide camera. This ensures that traffic lights, which may fall outside the FOV of the front-medium camera at close distances, are still visible. In such cases, the front-wide camera is able to see all traffic lights.

\subsection{Projection and Association}
\label{sec:Projection}
\subsubsection{Mapping} 
The detection model generates classified bounding boxes in the image space, assigning a confidence score to each prediction. These 2D bounding boxes must be associated with real-world traffic lights to enable informed driving decisions. We use an HD~map, where the position of traffic lights is stored in world coordinates calculated by the localization of the vehicle~\cite{ochs2025chefs}. The coordinates are either manually annotated by selecting the traffic lights in a localized point cloud during a mapping run or automatically calculated with a 3D-from-motion algorithm~\cite{fairfield2011traffic} if LiDAR data is unavailable. In addition to their positions, we also annotate the pictogram of each traffic light and create signal groups. A signal group is a collection of traffic lights that collectively manage a specific traffic movement. If more than one traffic light applies to a maneuver, like traveling straight ahead or turning right, these lights belong to the same signal group, ensuring they always display a consistent state together. For signal groups that support V2X technology and broadcast SPaT/MAP messages, we also store the Intersection and SignalPhase IDs.

\subsubsection{Projection}

The task of the projection and association module is to link the detected traffic lights in 2D image space with their corresponding positions in the HD map's 3D world space. The center point $(x_i, y_i)$ of each detected bounding box $d_i$ is used to create a ray $\mathbf{r}_i$ in 3D world coordinates, using Equation~\ref{eq:raycast}, with the intrinsic camera parameters $f_x, f_y, c_x, c_y$, the rotation matrix $\mathbf{R}$ which represents the transformation from the camera frame to the world frame, the unknown and variable distance scalar $t$, and the origin $\mathbf{o}$, which is the current position of the camera in the world coordinate system:
\vspace{-10pt}
\begin{equation}\label{eq:raycast}
\mathbf{r}_i = \mathbf{o} + t \cdot \mathbf{R} \cdot 
\frac{
\begin{bmatrix}
\frac{x_i - c_x}{f_x} \\
\frac{y_i - c_y}{f_y} \\
1
\end{bmatrix}
}{
\sqrt{\left(\frac{x_i - c_x}{f_x}\right)^2 + \left(\frac{y_i - c_y}{f_y}\right)^2 + 1}
}
\end{equation}

We define a region extending 180 meters in front of the vehicle. Within this region, the distance between every ray and every traffic light, represented by its 3D reference point $p_j$, is calculated. Assuming a perfect intrinsic and extrinsic camera calibration, vehicle localization, traffic light detection, and 3D traffic light mapping, every ray would intersect with a traffic light in 3D world space, and the association between a detection and a traffic light would be trivial. However, in practical applications, each of these components is subject to small errors.

\subsubsection{Association}
 To account for these imperfections, the association task is formulated as a weighted complete bipartite graph $G~=~(U, V, E)$.  In this graph, $U$ represents the set of 2D detections, each projected as a ray into the 3D world, while $V$ represents the set of mapped traffic lights' reference positions within a predefined region around the vehicle. The edges $E$ connect each node $u_i~\in~U$ to every node $v_j~\in~V$, forming a complete bipartite graph structure. 
The cost of each edge $e_{ij}~\in~E$ is defined as the geometric distance in meters between the ray $\mathbf{r}_i$ represented by node $u_i$ and the reference position $p_j$ of the corresponding mapped traffic light represented by node $v_j$. 
The edge costs are capped at a maximum value of 10 meters to constrain the optimization space and to prevent extremely high costs from dominating the optimization. 
To balance the bipartite graph for the minimum cost assignment problem, we augment the smaller node set with dummy nodes to ensure equal cardinality. The minimum cost matching is then solved using the Hungarian algorithm~\cite{hungarian}, resulting in an optimal one-to-one assignment $E_\text{assigned}~\subseteq~E$. A 2D detection $d_i$ with ray $\mathbf{r}_i$ is considered successfully associated with a traffic light $p_j$ if the assigned edge $e_{ij}~\in~E_\text{assigned}$ has a cost of less than 2 meters. 
\begin{figure}[tb]
\centering
\begin{subfigure}[htb]{\linewidth}
    \includegraphics[width=\textwidth]{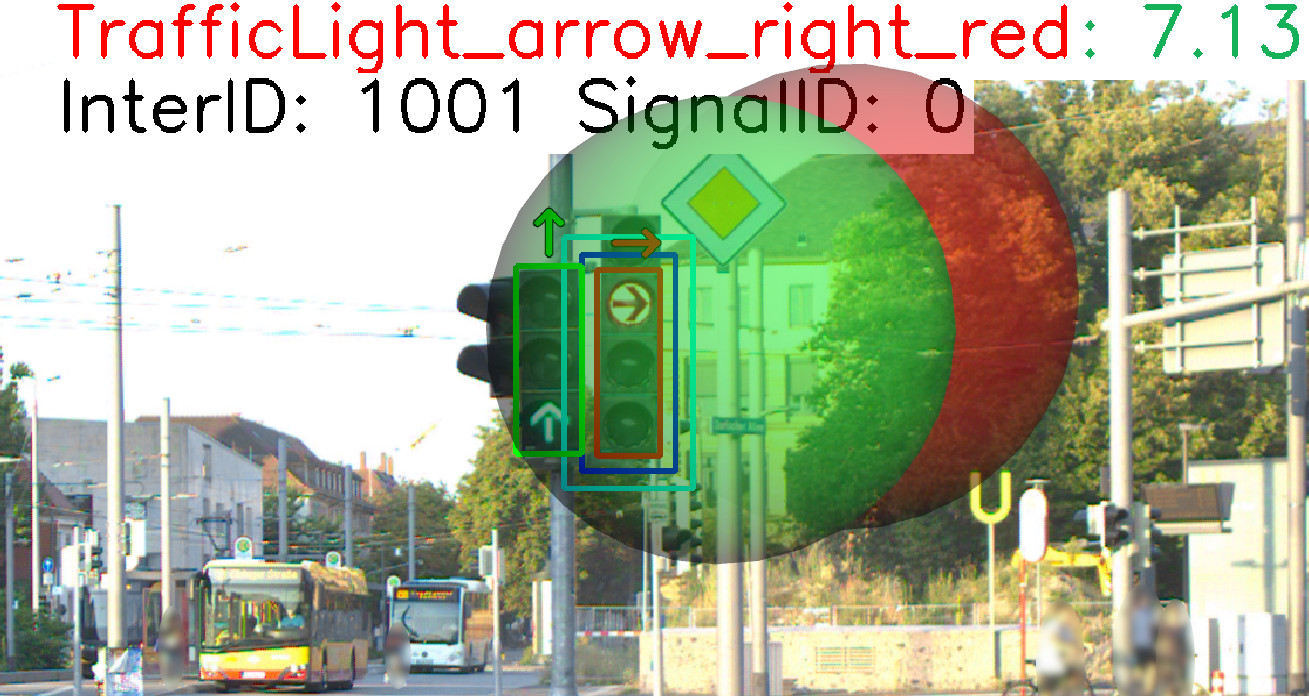}
\end{subfigure}
    \caption{CoCar NextGen, positioned in a right-turn lane, experiences a localization error that shifts the 3D projected traffic light positions (spheres) rightward. Despite this, our framework effectively minimizes the overall cost and correctly associates the red traffic light, as indicated by the blue and teal bounding boxes. The blue bounding box encompasses the relevant detection for the ego vehicle, and the teal bounding box shows high confidence. At the top, the decision for the autonomous driving stack and its color-coded confidence are displayed.}
    \label{fig:wrong_projection}
\end{figure}

Figure~\ref{fig:wrong_projection} illustrates the advantage of formulating the association problem as a graph and solving it using minimum cost matching. 
Without global optimization, a localization fault can introduce projection errors, causing the 2D detection on the right to be incorrectly matched to the left 3D projected traffic light. 
The regular approach, such as selecting the nearest traffic light~\cite{possatti2019traffic, fairfield2011traffic}, fails to account for these errors, leading to incorrect associations. 
By applying the Hungarian algorithm, the overall cost is minimized, ensuring that the detections are correctly associated with their corresponding traffic lights.

\subsubsection{Decision module}
Correct state estimations of traffic lights are critical for the autonomous driving stack. Therefore, we do not want to immediately assign a new state if we detect a change in states, as it could be an erroneous detection or assignment and demand multiple detections to change the state. To address this, we maintain a separate circular buffer for every traffic light, storing recent detections of both camera streams along with their timestamps, predicted classifications, and confidence scores. Each traffic light's final state is determined based on the detections stored in its respective buffer. The final state is computed as the one that maximizes the cumulative weight of all detections within the buffer. This makes sure that we are certain about our state and that some misdetections do not cause wrong states. We base the weight of each detection on the predicted confidence of the detector but adjust it in two ways. Firstly, we downweigh the confidence of a detection linearly based on the elapsed time in order to make older associated detections less relevant. We also drastically decrease the weight if the predicted pictogram of the detection does not match the pictogram of the traffic light in the HD map. This, in conjunction with the minimum cost matching association, reduces the likelihood that erroneous detections of other traffic lights change the state. Additionally, since all traffic lights within a signal group must display the same state and pictogram, we can reason about their associated detections. If a signal group has multiple traffic lights, we let the traffic light with the highest confidence in its state determine the state of the signal group, which in turn is the basis of the driving decision. In our experiments, a circular buffer size of 9, a linear weight decay to zero over three seconds, and halving the weight for mismatched pictograms yielded optimal decision module performance. These parameters need to be adjusted depending on the frequency, latency, and accuracy of the detection method and projection performance.
Our AD stack~\cite{ochs2024one} leverages the processed signal group states for informed decision-making and path planning. It first identifies which signal group is relevant to the planned route using the HD map and the vehicle’s trajectory. If the state of a relevant signal group is unknown, the system conservatively maps it to red. In the case of a yellow light, the planner calculates whether the vehicle can safely come to a halt before the stopping line. If stopping is feasible, the vehicle will decelerate and halt; otherwise, it will drive through the intersection. Additionally, if a SPaT/MAP V2X signal is available, the AD stack integrates this information in the decision-making process.

\begin{figure*}[t]

\begin{subfigure}[b]{\textwidth}
    \includegraphics[width=\textwidth]{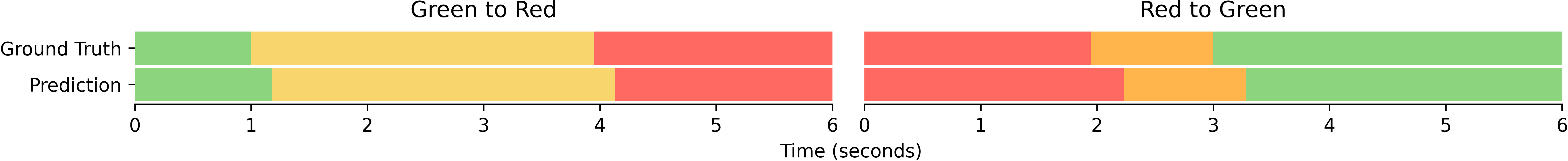}
    \captionsetup{labelformat=parens, labelsep=space}
    \caption*{Fig. 6a: Across our recordings, we observe an average end-to-end latency of 184~ms for signal group changes, primarily caused by the detector and the circular buffers in the decision module. The examples shown illustrate this effect.}
    \phantomsubcaption
    \label{fig:latency}
    \vspace{15pt}
\end{subfigure}

\begin{subfigure}[b]{\textwidth}
    \includegraphics[width=1\textwidth]{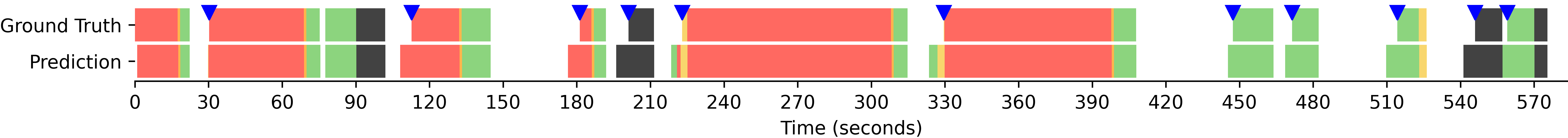}
    \captionsetup{labelformat=parens, labelsep=space}
    \caption*{Fig. 6b: Evaluation during the first 10 minutes of autonomous driving compares the ground truth and predicted signal phase state of the nearest relevant signal phase. A downward triangle indicates that a signal group's distance is 120 meters, the threshold at which labeling begins. Black indicates on-demand traffic lights that are currently \textit{off}. In most cases, our framework detects signals beyond 120 meters and nearly perfectly matches the ground truth. Only two minor erroneous state flips occur, both beyond the 120-meter range at 30 and 230 seconds.}
\phantomsubcaption
 \label{fig:10minute}

\end{subfigure}
\end{figure*}

\section{Evaluation}

Whenever possible, autonomous vehicles should avoid braking intensively for red lights to reduce the risk of rear-end collisions and enhance passenger comfort. In professional human-driven tests, we measure smooth braking maneuvers for distant red lights that are visible far ahead, with a deceleration rate of $-1~\text{m}/\text{s}^2$. The maximum urban speed limit in Germany is 50~km/h, which corresponds to a braking distance of 96 meters. Consequently, an autonomous driving system should aim to replicate this smooth braking behavior when red traffic lights are visible at or beyond this distance.

To evaluate the entire traffic light perception framework, we deploy it in CoCar NextGen~\cite{heinrich2024cocarnextgen}. For qualitative evaluation, we conducted 230 minutes of fully autonomous driving in the German city of Karlsruhe. During this time, the safety driver never needed to manually override the decision module, by either braking or accelerating, as the framework demonstrated very high reliability.
For a more fine-granular evaluation, we hand-annotate all 20~Hz camera streams for the first 10 minutes of this trip. Notably, this evaluation is conducted while driving fully autonomously, ensuring that the perception framework alone dictates vehicle speed adjustments. This prevents any human driver anticipation (e.g., braking or accelerating earlier) that could otherwise afford the system additional reaction time. For this evaluation, V2X messages are ignored, even when available~\cite{fleck2019towards}. We also annotate a 1-hour non-moving recording to better evaluate the latencies during state changes. 

We evaluate the reliability and robustness of the traffic light perception framework based on the following criteria, which are critical for ensuring safe and comfortable autonomous driving behavior:
\begin{enumerate}[label=\textit{\Alph*.}]
    \item Time delay between changes in a signal group state and its corresponding recognition.
    \item Stability of traffic light predictions and flickering.
    \item Perception accuracy of the module compared to a human-annotated ground truth.
    \item Minimum distance for reliable detection. 
\end{enumerate}

We analyze the recorded footage from three front-facing cameras, including their timestamps and distance to traffic lights. For each intersection, we select the first frame where the relevant traffic light is either 120~meters away or first comes into view. As the vehicle approaches the intersection, timestamps corresponding to traffic light color changes are logged. Our evaluation annotations are unaffected by human reaction time as we annotate individual frames but are limited by the 20~Hz image capture frequency of each camera.

First, we measure the latency between image capturing and the decision of our framework. Image preprocessing and networking take an average of 34~ms, the detector takes an average of 46~ms, and the association and decision modules take less than 1~ms to execute. This results in a latency between image capture and traffic light processing of 81~ms.
The weight maximization of the circular buffer in the decision module can negatively impact the latency when a signal group state changes, as the new detections with the updated state must generate enough weight to override previous state detections. On average, we measure a delay of 103~ms for state changes, as it takes between 3 to 5 detections to confirm and transition to the new state. This results in an average end-to-end reaction time of 184~ms and is exemplarily illustrated in Figure~\ref{fig:latency}. Our worst-case latency spike was 381~ms, which arose when the signal group consisted of only one traffic light which was hard to see and only captured by the wide-angle camera. Still, even in this case, it is faster than human drivers~\cite{green2000long}.

In a similar evaluation to ours, Possatti et al.~\cite{possatti2019traffic} observed their first correct detections at 84 meters but reported unstable model behavior, with rapid switching between predicted classes and even critical misclassifications, such as mistaking a \textit{red} traffic light for \textit{green}. Flickering leads to rapid acceleration and braking in autonomous driving stacks, and mistaking red traffic lights for green can cause severe accidents.
In contrast, we demonstrate that our framework is reliable, stable, and capable of detecting traffic lights from greater distances.
During the first 10-minute drive, depicted in Figure~\ref{fig:10minute}, no erroneous state changes were detected within a range of 120 meters. Two minor exceptions beyond this distance appeared under partial occlusions by a tree branch (30 seconds) and a post (220 seconds), causing brief misclassifications. Both situations involved turns governed by a single traffic light without the redundancy of multiple lights per signal group. Apart from these incidents, predictions were highly stable, with no flickering. Most of the stability is provided by the circular buffers, smoothing missed or wrong detections. Within the 120-meter range, our framework achieved a calculated accuracy of 99.33\%, whereas the missing 0.67\% are caused by latencies detailed in Figure~\ref{fig:latency}.

Figure~\ref{fig:10minute} also shows that in most cases, our model correctly predicts traffic light states beyond the 120-meter distance used for annotating the ground truth. The average distance of the first successful detection and association for traffic lights that are visible from a distance is 169.5~meters. These findings confirm that our framework accurately detects and classifies traffic signals with minimal latency, enabling smooth, human-like autonomous driving.

\newpage
\section{Conclusion}
In this paper, we presented a comprehensive approach to advancing camera-based traffic light detection and decision-making for autonomous vehicles. Our contributions included the creation of the \texttt{ATLAS} dataset, which remedies many of the key deficiencies in current public datasets by incorporating a wide range of traffic light states, pictograms, diverse environmental conditions, and a multi-camera setup with varying focal lengths. This dataset not only provides a richer training foundation but also enhances the generalization capabilities of detection models in real-world autonomous driving scenarios. By training state-of-the-art object detection models on \texttt{ATLAS}, we achieved significant improvements in detection accuracy and generalization across varied conditions. Furthermore, we introduced a robust association and decision-making framework that effectively integrates traffic light detections with mapped traffic light positions. This approach uses a bipartite graph-based matching strategy to address practical errors in calibration, localization, and detection necessary for real-world autonomous driving. Including a decision framework with a circular buffer mechanism further enhanced stability, ensuring that erroneous detections did not lead to flickering or unreliable state changes while keeping latencies low. Through extensive real-world testing, our perception framework demonstrated consistent accuracy and stability. In a range of scenarios, our approach reliably predicted relevant traffic light states and enabled safe autonomous navigation. 

\vspace{0.30cm}
\section*{Acknowledgment}

This work was supported by funding from the Topic Engineering Secure Systems of the Helmholtz
Association (HGF) and by KASTEL Security Research Labs (46.23.03).

{\small
\bibliographystyle{IEEEtran}
\bibliography{references}

\begin{thebibliography}{10}
\providecommand{\url}[1]{#1}
\csname url@samestyle\endcsname
\providecommand{\newblock}{\relax}
\providecommand{\bibinfo}[2]{#2}
\providecommand{\BIBentrySTDinterwordspacing}{\spaceskip=0pt\relax}
\providecommand{\BIBentryALTinterwordstretchfactor}{4}
\providecommand{\BIBentryALTinterwordspacing}{\spaceskip=\fontdimen2\font plus
\BIBentryALTinterwordstretchfactor\fontdimen3\font minus \fontdimen4\font\relax}
\providecommand{\BIBforeignlanguage}[2]{{%
\expandafter\ifx\csname l@#1\endcsname\relax
\typeout{** WARNING: IEEEtran.bst: No hyphenation pattern has been}%
\typeout{** loaded for the language `#1'. Using the pattern for}%
\typeout{** the default language instead.}%
\else
\language=\csname l@#1\endcsname
\fi
#2}}
\providecommand{\BIBdecl}{\relax}
\BIBdecl

\bibitem{chung2002vision}
Y.-C. Chung, J.-M. Wang, and S.-W. Chen, ``{A Vision-Based Traffic Light Detection System at Intersections},'' \emph{Journal of Taiwan Normal University: Mathematics, Science and Technology}, 2002.

\bibitem{de2009real}
R.~De~Charette and F.~Nashashibi, ``{Real time visual traffic lights recognition based on Spot Light Detection and adaptive traffic lights templates},'' in \emph{Intelligent Vehicles Symposium (IV)}.\hskip 1em plus 0.5em minus 0.4em\relax IEEE, 2009.

\bibitem{lindner2004robust}
F.~Lindner, U.~Kressel, and S.~Kaelberer, ``Robust recognition of traffic signals,'' in \emph{Intelligent Vehicles Symposium (IV)}.\hskip 1em plus 0.5em minus 0.4em\relax IEEE, 2004.

\bibitem{nienhuser2010visual}
D.~Nienh{\"u}ser, M.~Drescher, and J.~M. Z{\"o}llner, ``{Visual state estimation of traffic lights using hidden Markov models},'' in \emph{International Conference on Intelligent Transportation Systems (ITSC)}.\hskip 1em plus 0.5em minus 0.4em\relax IEEE, 2010.

\bibitem{pavlitska2023traffic}
S.~Pavlitska, N.~Lambing, A.~K. Bangaru, and J.~M. Z{\"o}llner, ``{Traffic Light Recognition using Convolutional Neural Networks: A Survey},'' in \emph{International Conference on Intelligent Transportation Systems (ITSC)}, 2023.

\bibitem{possatti2019traffic}
{L. C. Possatti, R. Guidolini, V. Cardoso, et al.}, ``{Traffic Light Recognition Using Deep Learning and Prior Maps for Autonomous Cars},'' in \emph{International Joint Conference on Neural Networks (IJCNN)}.\hskip 1em plus 0.5em minus 0.4em\relax IEEE, 2019.

\bibitem{chen2024traffic}
Y.-C. Chen and H.-Y. Lin, ``{Traffic Light Detection and Recognition using Ensemble Learning with Color-Based Data Augmentation},'' in \emph{Intelligent Vehicles Symposium (IV)}.\hskip 1em plus 0.5em minus 0.4em\relax IEEE, 2024.

\bibitem{jocher2023yolov8}
G.~Jocher, A.~Chaurasia, and J.~Qiu, ``{YOLO by Ultralytics (Version 8.0.0) [Computer software]},'' \url{https://github.com/ultralytics/ultralytics}, 2023.

\bibitem{tldready}
N.~Polley, S.~Pavlitska, Y.~Boualili, P.~Rohrbeck, P.~Stiller, A.~K. Bangaru, and J.~M. Z{\"o}llner, ``{TLD-READY: Traffic Light Detection - Relevance Estimation and Deployment Analysis},'' in \emph{International Conference on Intelligent Transportation Systems (ITSC)}, 2024.

\bibitem{Geiger2012CVPR}
A.~Geiger, P.~Lenz, and R.~Urtasun, ``{Are we ready for Autonomous Driving? The KITTI Vision Benchmark Suite},'' in \emph{Conference on Computer Vision and Pattern Recognition (CVPR)}, 2012.

\bibitem{nuscenes}
H.~Caesar, V.~Bankiti, A.~H. Lang, S.~Vora, V.~E. Liong, Q.~Xu, A.~Krishnan, Y.~Pan, G.~Baldan, and O.~Beijbom, ``{nuScenes: A multimodal dataset for autonomous driving},'' in \emph{Conference on Computer Vision and Pattern Recognition (CVPR)}, 2020.

\bibitem{LARA}
R.~de~Charette, ``{LaRA french traffic lights recognition (tlr) public benchmarks},'' 2015.

\bibitem{kunsagi2024aimotive}
S.~Kuns{\'a}gi-M{\'a}t{\'e}, L.~Pet{\H{o}}, L.~Seres, and T.~Matuszka, ``{Accurate Automatic 3D Annotation of Traffic Lights and Signs for Autonomous Driving},'' \emph{European Conference on Computer Vision 2024 Workshop on Vision-Centric Autonomous Driving}.

\bibitem{behrendt2017deep}
K.~Behrendt, L.~Novak, and R.~Botros, ``{A deep learning approach to traffic lights: Detection, tracking, and classification},'' in \emph{International Conference on Robotics and Automation (ICRA)}.\hskip 1em plus 0.5em minus 0.4em\relax IEEE, 2017.

\bibitem{jensen2016vision}
M.~B. Jensen, M.~P. Philipsen, A.~M{\o}gelmose, T.~B. Moeslund, and M.~M. Trivedi, ``{Vision for Looking at Traffic Lights: Issues, Survey, and Perspectives},'' \emph{IEEE Transactions on Intelligent Transportation Systems}, 2016.

\bibitem{fregin2018driveu}
A.~Fregin, J.~Mueller, U.~Kreßel, and K.~Dietmayer, ``{The DriveU traffic light dataset: Introduction and comparison with existing datasets},'' in \emph{International Conference on Robotics and Automation (ICRA)}.\hskip 1em plus 0.5em minus 0.4em\relax IEEE, 2018.

\bibitem{li2017traffic}
X.~Li, H.~Ma, X.~Wang, and X.~Zhang, ``{Traffic Light Recognition for Complex Scene With Fusion Detections},'' \emph{International Conference on Intelligent Transportation Systems (ITSC)}, 2017.

\bibitem{langenberg2019deep}
T.~Langenberg, T.~L{\"u}ddecke, and F.~W{\"o}rg{\"o}tter, ``{Deep Metadata Fusion for Traffic Light to Lane Assignment},'' \emph{IEEE Robotics and Automation Letters}, 2019.

\bibitem{fairfield2011traffic}
N.~Fairfield and C.~Urmson, ``Traffic light mapping and detection,'' in \emph{International Conference on Robotics and Automation (ICRA)}.\hskip 1em plus 0.5em minus 0.4em\relax IEEE, 2011.

\bibitem{heinrich2024cocarnextgen}
M.~Heinrich, M.~Zipfl, M.~Uecker, S.~Ochs, M.~Gontscharow, T.~Fleck, J.~Doll, P.~Schörner, C.~Hubschneider, M.~R. Zofka, A.~Viehl, and J.~M. Zöllner, ``{CoCar NextGen: A Multi-Purpose Platform for Connected Autonomous Driving Research},'' in \emph{International Conference on Intelligent Transportation Systems (ITSC)}, 2024.

\bibitem{hukkelas23DP2}
H.~Hukkelås and F.~Lindseth, ``{DeepPrivacy2: Towards Realistic Full-Body Anonymization},'' in \emph{Winter Conference on Applications of Computer Vision (WACV)}, 2023.

\bibitem{raina2023egoblur}
{N. Raina, G. Somasundaram, K. Zheng, S. Miglani, S. Saarinen, J.~Meissner, M. Schwesinger et al.}, ``{EgoBlur: Responsible Innovation in Aria},'' \emph{arXiv preprint arXiv:2308.13093}, 2023.

\bibitem{wang2025yolov9}
C.-Y. Wang, I.-H. Yeh, and H.-Y. Mark~Liao, ``{YOLOv9: Learning What You Want to Learn Using Programmable Gradient Information},'' in \emph{European Conference on Computer Vision (ECCV)}.\hskip 1em plus 0.5em minus 0.4em\relax Springer, 2025.

\bibitem{wang2024yolov10}
A.~Wang, H.~Chen, L.~Liu, K.~Chen, Z.~Lin, J.~Han, and G.~Ding, ``{YOLOv10: Real-Time End-to-End Object Detection},'' in \emph{Advances in Neural Information Processing Systems (NIPS)}, 2024.

\bibitem{yolo11_ultralytics}
\BIBentryALTinterwordspacing
G.~Jocher and J.~Qiu, ``{Ultralytics YOLO11},'' 2025. [Online]. Available: \url{https://github.com/ultralytics/ultralytics}
\BIBentrySTDinterwordspacing

\bibitem{tian2025yolov12}
Y.~Tian, Q.~Ye, and D.~Doermann, ``{YOLOv12: Attention-Centric Real-Time Object Detectors},'' \emph{arXiv preprint arXiv:2502.12524}, 2025.

\bibitem{ochs2025chefs}
S.~Ochs, M.~Heinrich, P.~Sch{\"o}rner, M.~R. Zofka, and J.~M. Z{\"o}llner, ``{A Chefs KISS--Utilizing semantic information in both ICP and SLAM framework},'' \emph{arXiv preprint arXiv:2504.02086}, 2025.

\bibitem{hungarian}
H.~W. Kuhn, ``{The Hungarian method for the assignment problem},'' \emph{Naval research logistics quarterly}, 1955.

\bibitem{ochs2024one}
S.~Ochs, J.~Doll, D.~Grimm, T.~Fleck, M.~Heinrich, S.~Orf, A.~Schotschneider, H.~Gremmelmaier, R.~Polley, S.~Pavlitska \emph{et~al.}, ``{One Stack to Rule them All: To Drive Automated Vehicles, and Reach for the 4th level},'' \emph{arXiv preprint arXiv:2404.02645}, 2024.

\bibitem{fleck2019towards}
T.~Fleck, K.~Daaboul, M.~Weber, P.~Sch{\"o}rner, M.~Wehmer, J.~Doll, S.~Orf, N.~Su{\ss}mann, C.~Hubschneider, M.~R. Zofka \emph{et~al.}, ``{Towards Large Scale Urban Traffic Reference Data: Smart Infrastructure in the Test Area Autonomous Driving Baden-W{\"u}rttemberg},'' in \emph{Intelligent Autonomous Systems 15}.\hskip 1em plus 0.5em minus 0.4em\relax Springer, 2019.

\bibitem{green2000long}
M.~Green, ``{"How long does it take to stop?" Methodological analysis of driver perception-brake times},'' \emph{Transportation Human Factors, Volume 2, Issue 3}, 2000.

\end{thebibliography}
}

\end{document}